\documentclass[11pt,a4paper]{article}
\usepackage[hyperref]{acl2021}
\usepackage{times}
\usepackage{latexsym}

\usepackage{adjustbox}
\usepackage{bm}

\usepackage{booktabs}
\usepackage{multirow}
\usepackage{array}
\usepackage{adjustbox}
\usepackage{enumitem}
\usepackage{hhline}

\usepackage{microtype}
\usepackage{amsmath}
\usepackage{bbm}
\aclfinalcopy 

\title{Multimodal Graph-based Transformer Framework for Biomedical Relation Extraction}

\author{
Sriram Pingali, \hspace{0.1cm} Shweta Yadav\thanks{$^*$These authors contributed equally to this work. \phantom{--------} Accepted for publication at the Joint Conference of the 59th Annual Meeting of the Association for Computational Linguistics and the 11th International Joint Conference on Natural Language Processing (ACL-IJCNLP 2021)}, \hspace{0.1cm} Pratik Dutta\footnotemark[1], \hspace{0.1cm} Sriparna Saha  \\
   Indian Institute of Technology Patna, India  \\
  {\tt 
  \{1801cs37}, 
    {\tt 
  shweta.pcs14},
      {\tt 
  pratik.pcs16},
{\tt 
  sriparna\}@iitp.ac.in}
  } 

\date{}

\begin{document}
\maketitle
\begin{abstract}
The recent advancement of pre-trained Transformer models has propelled the development of effective text mining models across various biomedical tasks. However, these models are primarily learned on the textual data and often lack the domain knowledge of the entities to capture the context beyond the sentence. 
In this study, we introduced a novel framework that enables the model to learn multi-omnics biological information about entities (proteins) with the help of additional multi-modal cues like molecular structure.
Towards this, rather developing modality-specific architectures, we devise a generalized and optimized graph based multi-modal learning mechanism that utilizes the GraphBERT model to encode the textual and molecular structure information and exploit the underlying features of various modalities to enable the end-to-end learning. 
We evaluated our proposed method on Protein-Protein Interaction task from the biomedical corpus, where our proposed generalized approach is observed to be benefited by the additional domain-specific modality. 

\end{abstract}

\section{Introduction}
The biomedical scientific articles hold the valuable knowledge of biomedical entities (such as protein, drug, gene) and their relationships. However, with the exponential increase in the volume of biomedical articles \cite{lu2011pubmed}, it is imperative to advance the development of an accurate biomedical text mining tool to extract and curate meaningful information from huge unstructured texts automatically.\\
\indent One of the cardinal tasks in biomedical document processing is Protein-protein interaction (PPI), where the relation (`\textit{interaction}' or `\textit{non-interaction}') between  two protein mentions is identified from the given biomedical text. The knowledge about protein interactions is critical in understanding the biological processes, such as signaling cascades, translations and metabolism, that are regulated by the interactions of proteins that alter proteins to modulate their stability \cite{elangovan2020assigning}. 

Majority of the existing works on PPI in the literature primarily focused only on the textual information present in the biomedical article. However, these approaches lack in capturing \textbf{(1)} multi-omnics biological information regarding protein interactions, and \textbf{(2)} genetic and structure information of the proteins. A few works \cite{dutta-saha-2020-amalgamation,asada-etal-2018-enhancing, DBLP:conf/iconip/JhaSK20, Jha2020} have been reported in the literature where the researchers have considered different modalities of the biomedical corpus. 
 However, these multi-modal architectures are modality-specific and thus are very complex. Hence, there is a surge to develop a generalized and optimized model that can understand all the modalities rather than developing various architectures for different modalities.  \\
\indent Towards this, we explore Graph-based Transformer model (GraphBERT) \cite{zhang2020graph} to learn the modality independent graph representation. This enables the model to acquire the joint knowledge of both the modalities (textual and protein structure) under a single learning network. 
The main contributions of this work are:
\begin{enumerate}[noitemsep]
    \item Besides the textual information of the biomedical corpus, we have also utilized protein atomic structural information while identifying the protein interactions.
    \item Developed a generalized modality-agnostic approach that is able to learn the feature representations of both the textual and the protein-structural modality.
    \item Our analysis reveals that addition of protein-structure modality increases the efficiency of model in identifying the interacted protein mentions.
\end{enumerate}
\paragraph{Related Work:}\label{related}
Existing studies have adopted traditional statistical and graphical methods \cite{miyao2008evaluating,chang2016pipe} to identify the protein interactions from the textual content. 
Later, with the success of deep learning, several techniques based on Convolutional Neural Network \cite{choi2018extraction,peng2017deep,ekbal2016deep}, Recurrent Neural Network \cite{hsieh2017identifying,ahmed2019identifying}, Long Short Term Memory network \cite{yadav2019feature,ningthoujam2019relation,yadav2020relation}, and language models \cite{yadav2021they} based methods are proposed for extracting the relationships from biomedical literature and clinical records.
\citet{fei2020span} proposed a span-graph neural model for jointly extracting overlapping entity relationships from biomedical text. 
The recent advancement of the Transformer model \cite{lee2020biobert,beltagy2019scibert} in the biomedical domain has also led to significant performance improvement in biomedical relation extraction task \cite{giles2020optimising}. 
Recently, the use of multi-modal dataset in BioNLP domain \cite{dutta-saha-2020-amalgamation, asada-etal-2018-enhancing} draws the attention of the researchers due to its better performance than the traditional approaches. In contrast, our model is independent of handling multiple modalities without relying on modality-specific architectures.  

\section{Proposed Method}\label{proposed}
\begin{figure*}
\centering
\includegraphics[width=\textwidth]{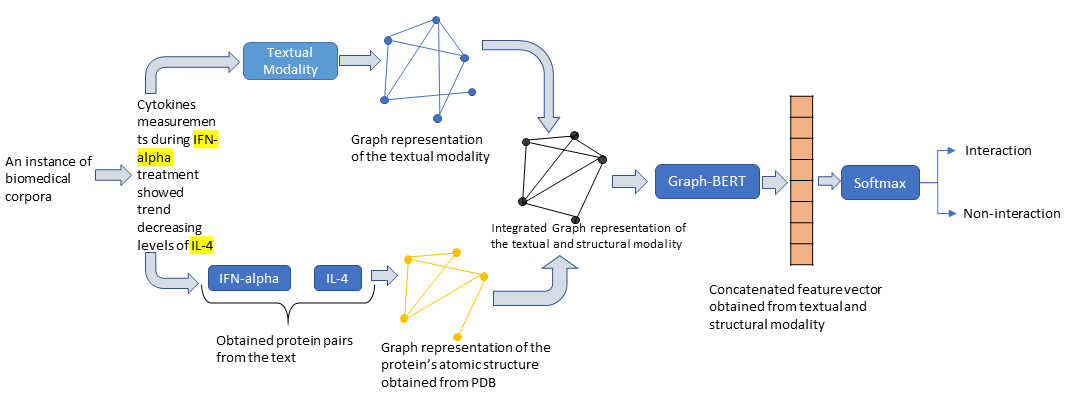}
\caption{An outline of the proposed deep multi-modal architecture for PPI.}
\label{figure-1}
\end{figure*}

In this section, we introduce our proposed method and its detailed implementation.
The proposed deep multi-modal architecture is illustrated in Figure-\ref{figure-1}, that consists of four main components:
\textbf{(1)} \textit{Multi-modal Graph Constructor}, \textbf{(2)} \textit{Multi-modal Graph Fusion}, \textbf{(3)} \textit{Multi-modal Graph Encoder}, \textbf{(4)} \textit{PPI Predictor}.
Below we briefly describe each of the model components.
\paragraph{Problem Statement:} Given a biomedical input text $S=\{w_1, w_2, \ldots, w_n\}$ having $n$ words, and a pair of protein mentions $p_1, p_2 \in S$, we aim to predict, whether the protein mentions will \textit{`interact'} or \textit{`non-interact'}.
\subsection{Multi-modal Graph Constructor} This component consists of two distinct graph constructors for two different modalities, which are Textual Graph Constructor and Protein Structure Graph Constructor. The former, constructs the graph by considering the textual content that aims to capture the lexical and contextual information present in the input. The later, exploits the atomic structure (3D PDB structure) of the protein molecules to build the graph.
\paragraph{Textual Graph Constructor:} 
To generate the textual graph, we begin by first constructing the vocabulary from the training corpus. For each input text $S$, we use one-hot-encoding mechanism to encode them as a vector representation $R_S \in \mathcal{R}^{|V|}$. However, the representation $R_S$ suffers from the data sparsity as the vocabulary size can become very large for the entire training corpus. To deal with this, we utilized the Principal Component Analysis (PCA) \cite{wold1987principal} to reduce the vector dimensionality. The textual graph $\mathcal{G}_T =\{\mathcal{V}_T, \mathcal{E}_T \}$ is formulated by the nodes $\mathcal{V}_T=\{ \hat{R}_{S_1}, \hat{R}_{S_2}, \ldots, \hat{R}_{S_{|N|}}\}$, where $|N|$ is the number of input sentences in the training dataset and $\hat{R}_{S_i} \in \mathcal{R}^{|\hat{V}|}$ is reduced vector representation of size $|\hat{V}|$ for sentence $S_i$. The link $e_{i,j}$ between nodes $\hat{R}_{S_i}$ and $\hat{R}_{S_j}$ is determined by the common entities (\textit{protein}) present in both the sentences $S_i$ and $S_j$, if there is no common entity, then link does not exist between the nodes. The edges $\mathcal{E}_T = \{ e_{i, j} \ | \ i, j \in \mathcal{V}_T, \ \text{and} \ \text{\textit{protein}} \in i, \ j \}$ are the set of all the links that exist between any two nodes in the graph, $\mathcal{G}_T$.


\paragraph{Protein Structure Graph Constructor:}
For the protein structural modality, we created a graph where each node represents an atom and the edge represents the connection between the atoms. To obtain the atomic information about the proteins, first we have mapped the proteins into genes and utilized the PDB (Protein Data Bank)\footnote{https://www.rcsb.org/} for each associated protein mention.
Each protein information obtained from PDB consists of set of atoms $\{a_1, a_2, \ldots,  a_A\}$, and a
node feature matrix, $N_p \in \mathcal{R}^{A \times d_p}$. The node feature matrix for each protein $k$ undergoes the convolutional operation \textit{CNN(.)} followed by the max-pooling operation, $pool(.)$. Formally, $P_{k}=pool(relu(\text{\textit{CNN}}(N_{p_k})))$. The final protein representation, $P_{S_i}$, for both the proteins present in the given input sentence $S_i$ is computed as follows: $P_{S_i} = P_1 \oplus P_2$. Following this, the protein structure graph $\mathcal{G}_P =\{\mathcal{V}_P, \mathcal{E}_P \}$ is formulated by the nodes $\mathcal{V}_P=\{ {P}_{S_1}, {P}_{S_2}, \ldots, {P}_{S_{|N|}}\}$, where $|N|$ is the number of input sentences in the training dataset and ${P}_{S_i} \in \mathcal{R}^{d_s}$ is the protein structure representation of size $d_s$ for sentence $S_i$.

\subsection{Multi-modal Graph Fusion} 
In this component, we fused the textual graph $\mathcal{G}_T$ and protein structure graph $\mathcal{G}_P$ with the aim of generating a joint representation that is capable of capturing the contextual, lexical, and multi-omnics information. Towards this, we expanded the node information of textual graph with the node information obtained in the protein-structure graph. Specifically, we created a multi-modal graph $\mathcal{G}$ with the nodes $\mathcal{V}$ having concatenated vector representations from the respective nodes of textual graph and protein structure graph. Formally,
\begin{equation}
    \mathcal{V}_i = \hat{R}_{S_i} \oplus {P}_{S_i}
\end{equation}
The link information remains intact in the multi-modal graph fusion, thus, $\mathcal{E} = \mathcal{E_T}$.



\subsection{Multi-modal Graph Encoder} Majority of the existing works on multi-modal relation extraction have treated multiple modalities separately and exploited the modality-specific architectures to learn the respective feature representations. However, these strategies inhibit the learning of inherent shared complementary features, that are often present across the modalities. 
To address this, we present an end-to-end multi-modality learning mechanism that exploits the single expanded multi-modal graph (obtained from the Multi-modal Graph Expansion component) with the Graph-based Transformer encoder. Specifically, we utilized the Graph-BERT \cite{zhang2020graph} encoder over the other dominants graph neural networks (GNNs) primarily due to its capability to avoid the \textbf{(a)} \textit{suspended animation problem} \cite{zhang2019gresnet}, and \textbf{(b)} \textit{over-smoothing problem} \cite{li2018deeper} that hinders the applications of GNNs for deep graph representation learning tasks. 
For a given multi-modal graph $\mathcal{G}=(\mathcal{V}, \mathcal{E})$ with the set of nodes ($\mathcal{V}$) and edges ($\mathcal{E}$), Graph-BERT sampled set of graph batches for all the nodes as set $\mathcal{G} = \{g_1, g_2, \ldots, g_{|\mathcal{V}|}\}$. For all the nodes $v_j$ in sub-graph $g_i$, the Graph-BERT computes raw feature vector embedding $e_j^{x}$, role embedding $e_j^{r}$, position embedding $e_j^{p}$ and distance embedding $e_j^{d}$. The initial input vector for node $v_j$ is computed as follows: $h_{j}^{(0)}= e_j^{x} + e_j^{r} + e_j^{p} +e_j^{d}$. Furthermore, the initial input vectors for all the nodes in $g_i$ can be organized into a matrix $H^{(0)} =
[h_i^{(0)}, h_{i,1}^{(0)}, \ldots, h_{i,k}^{(0)}]^\top$, where $k$ is a hyper-parameter. The Graph-Transformer \cite{zhang2020graph} computes the vector representation of $D$ layers of transformers. The final feature ($z_i$) for node $v_j$ is computed as follows:
\begin{equation}
\begin{aligned}
{H}^{(0)} &  = [{h}_i^{(0)}, {h}_{i,1}^{(0)}, \cdots, {h}_{i,k}^{(0)}]^\top \\
{H}^{(l)} &= \text{\textit{G-Transformer}} \left( {H}^{(l-1)}\right)\\
{z}_i &= \sum_{m=0}^{m=D} {H}^{(m)}
\end{aligned}
\end{equation}

\subsection{PPI Predictor}
The final feature ($z_i$) of each node $i$ is used to predict  the PPI category. Towards this, we employed a feed-forward network with \textit{softmax} activation layer to predict the input text into one of the two classes \textit{interaction} or \textit{non-interaction}. Formally,
\begin{equation}
\footnotesize
    prob ({c=interact}|\mathcal{G}, S, \theta) = \frac{exp(\bm{W}^{T}z_i^{\text{interact}}+\bm{b})}{\sum_{k=1}^{K}{exp(\bm{W}^{T}z_i^{k}+\bm{b})}} 
\end{equation}
where, $\bm{W}$ and $\bm{b}$ are the weight matrix and bias vector, respectively. $K$ denotes total number of distinct classes, which are \textit{`interaction'} and \textit{`non-interaction'} in our case.

\begin{table}[t]
\begin{adjustbox}{max width=\linewidth}
\begin{tabular}{@{}lllcccl@{}}
\toprule \toprule
 &  &  & Precision & Recall & F-score &  \\ \midrule
 & \textbf{Proposed Model} &  & 80.84 & {80.87} & 80.86 &  \\
 & \citet{dutta-saha-2020-amalgamation} &  & 69.04 & 88.49 & 77.54 &  \\
 & \citet{yadav2019feature} &  & 80.81 & 82.57 & 81.68 &  \\
 & \citet{hua2016shortest} &  & 73.40 & 77.00 & 75.20 &  \\
 & \citet{choi2018extraction} &  & 72.05 & 77.51 & 74.68 &  \\
 & \citet{qian2012tree}  & & 63.61 & 61.24 & 62.40 &  \\
 &  \citet{peng2017deep} &  & 62.70 & 68.2 & 65.30 &  \\
 & \citet{zhao2016protein} &  & 53.90 & 72.9 & 61.60 &  \\
 & \citet{tikk2010comprehensive} &  & 53.30 & 70.10 & 60.00 &  \\
 & \citet{li2015approach} &  & 72.33 & 74.94 & 73.61 &  \\
 & \citet{choi2010simplicity} &  & 74.50 & 70.90 & 72.60 &  \\
 \bottomrule \bottomrule
\end{tabular}
\end{adjustbox}
\caption{Comparative analysis of the proposed multi-modal approach with state-of-the-art techniques for BioInfer dataset.}\label{table:2}
\vspace{-2mm}
\end{table}
\vspace{-2mm}
\begin{table}[t]
\begin{adjustbox}{max width=\linewidth}
\begin{tabular}{@{}clccc@{}}
\toprule
 &  & Precision & Recall & F-score \\ \midrule
\multirow{2}{*}{HPRD50} & Textual Modality & 90.44 & 92.18 & 91.28 \\
 & Proposed Model & 95.47 & 94.69 & 95.06 \\ \midrule
\multirow{2}{*}{BioInfer} & Textual Modality & 78.49 & 79.78 & 79.06 \\
 & Proposed Model & 80.84 & 80.87 & 80.86 \\ \bottomrule
\end{tabular}
\end{adjustbox}
\caption{Results by uni-modal and multimodal approaches}\label{table:1}
\end{table}
\section{Datasets and Experimental Analysis}\label{result}
{\bf Datasets:}
In this work, we have collected two exemplified multi-modal protein protein interaction datasets \cite{dutta-saha-2020-amalgamation}. In these datasets, the authors exemplified two popular benchmark PPI corpora, namely BioInfer\footnote{http://corpora.informatik.hu-berlin.de/} and HPRD50\footnote{https://goo.gl/M5tEJj}.

\paragraph{Experimental Setup}
We have utilized the pre-trained Graph-BERT\footnote{https://github.com/jwzhanggy/Graph-Bert} in our experiment.
The initial vocabulary for BioInfer and HPRD50 datasets are 6561 and 1277, respectively. We have projected them into 1000 and 1185 dimension vectors using PCA, respectively. We have kept maximum of 5052 and 1185 number of words in both the datasets, respectively. The filter-size of CNN is set to $3,4$. We  have obtained $1185$ length node feature representation for protein structure graph. The nodes of multi-modal graph received the $2185$ sized feature representation. We have obtained $2500$ and $25859$ number of nodes and edges from HPRD50 dataset and 13675 and 15930214 number of nodes and edges from BioInfer dataset for the Graph-BERT training, respectively. We have used all the hyper-parameters of Graph-BERT model in our proposed model. We have kept following hyper-parameters values: subgraph size = 5, hidden size = 32, attention head number = 2, Transformer layers, D = 2, learning rate = 0.01, weight decay = 5e-4, hidden dropout rate = 0.5, attention dropout rate = 0.3, loss = cross entropy, optimizer = adam \cite{adam}. The hyper-parameters are chosen based on the 5-fold cross-validation experiments on both the datasets.

{\bf Results and Analysis:}
We have compared the performance (\textit{c.f.} Table-\ref{table:2},\ref{table:3}) of our proposed model with the existing state-of-the-art methods on PPI for both the datasets. These existing methods are based on different techniques like kernel-based \cite{choi2010simplicity, tikk2010comprehensive, qian2012tree, li2015approach}, deep neural network-based \cite{zhao2016protein, yadav2019feature}, multi-channel dependency-based convolutional neural network model \cite{peng2017deep}, semantic feature embedding \cite{choi2018extraction}, shortest dependency path \cite{hua2016shortest} and a recent deep multi-modal approach \cite{dutta-saha-2020-amalgamation}. It is to be noted that our results on BioInfer and HPRD50 are not directly comparable with the existing approaches as other methods have utilized different test sets for evaluation.
From the above comparative study, it is evident that our proposed multi-modal approach identifies the protein interactions in an efficient way and can be further improved in different ways.

\begin{table}[t]
\begin{adjustbox}{max width=\linewidth}
\begin{tabular}{@{}lllcccl@{}}
\toprule \toprule
&  &  & Precision & Recall & F-score &  \\ \midrule
 & \textbf{Proposed Model} &  & 95.47 & 94.69 & 95.06 &  \\
  & \citet{dutta-saha-2020-amalgamation} &  & 94.79 & 75.21 & 83.87 &  \\
 & \citet{yadav2019feature} &  & 79.92 & 77.58 & 78.73 &  \\
 & \citet{tikk2010comprehensive} &  & 68.20 & 69.80 & 67.80 &  \\
 & \citet{tikk2010comprehensive}(with SVM) &  & 68.20 & 69.80 & 67.80 &  \\
 & \citet{palaga2009extracting} &  & 66.70 & 80.20 & 70.90 &  \\
 & \citet{airola2008all}(APG) &  & 64.30 & 65.80 & 63.40 &  \\
 & \citet{van2008extracting} &  & 60.00 & 51.00 & 55.00 &  \\
 & \citet{miwa2009protein} &  & 68.50 & 76.10 & 70.90 &  \\
 & \citet{airola2008all}(Co-occ) &  & 38.90 & 100 & 55.40 &  \\
 & \citet{pyysalo2008comparative} &  & 76.00 & 64.00 & 69.00 &  \\
 \bottomrule \bottomrule
\end{tabular}
\end{adjustbox}
\caption{Comparative analysis of the proposed multi-modal approach with other state-of-the-art approaches for HPRD50 dataset.}\label{table:3}
\end{table}

\begin{figure*}[!t]
\centering
\includegraphics[width=0.70\textwidth]{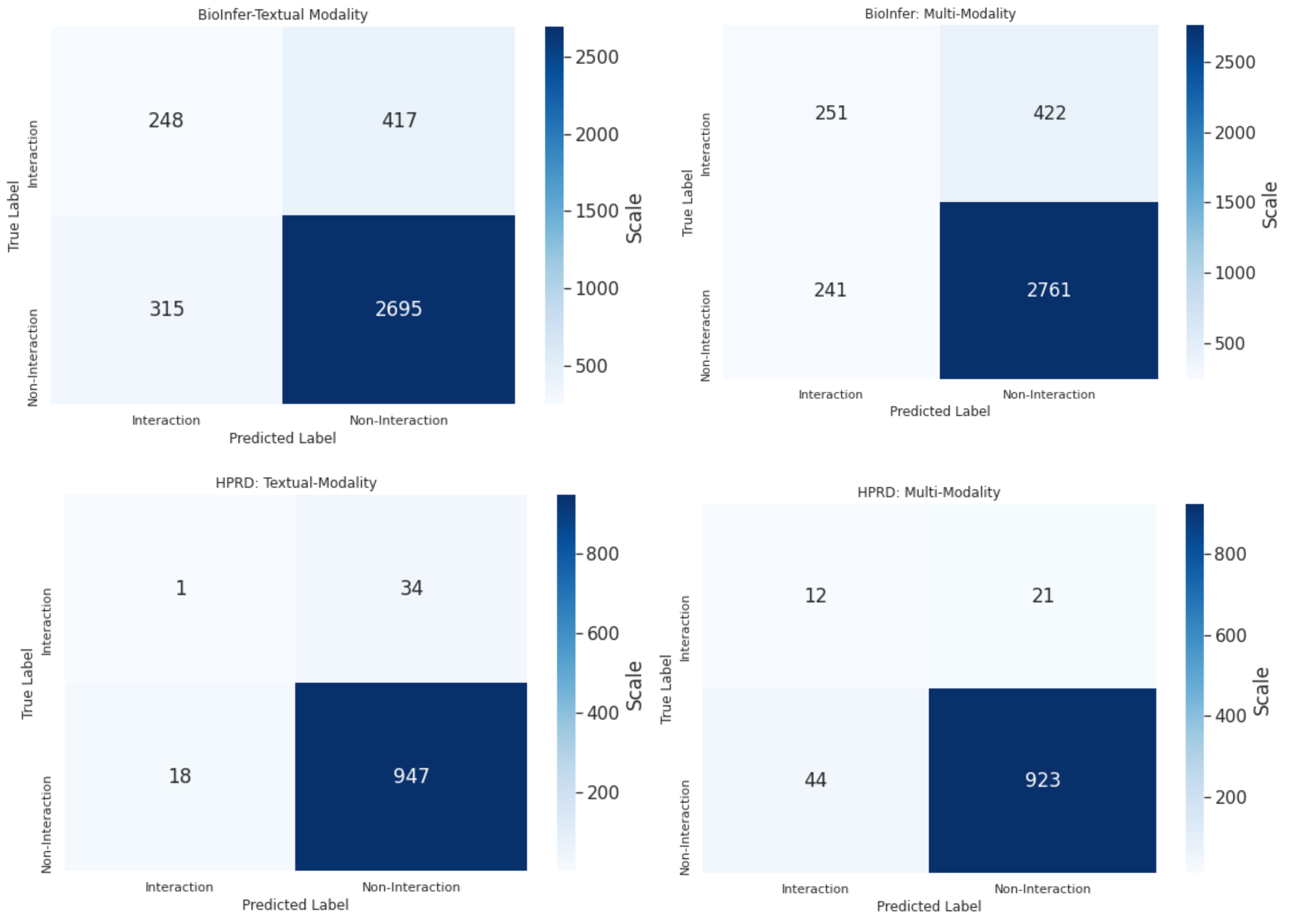}
\caption{Confusion matrices of our proposed approach on both HPRD and BioInfer datasets with only textual modality and text+structure modality.}\label{figure-2}
\end{figure*}
\noindent \textbf{Discussion:}
To analyze the role of each modality, we conducted ablation study as shown in Table \ref{table:1}.  We performed the experiments with the textual modality. Here, we could not consider the protein-structural modality alone as it would bring the conflicting labeling relation. For example, consider two sentences that contain same pair of proteins but these proteins can have conflicting relations (interacting or non-interacting) depending on the context of sentences in which they appear. Hence, we could not consider the protein-structural modality alone. Though the structural modality is unable to draw any conclusion alone, however the integration of both the modalities demonstrates the improvements (3.78\% and 1.8\%, in terms of F-score for HPRD50 and BioInfer, respectively) over the textual modality alone.

\section{Error Analysis}
The comparative confusion matrices with only textual-modality and multi-modality for both the datasets are shown in Figure-\ref{figure-2}. We have performed error analysis to postulate possible reasons and areas with scope of improvement in our experiments. After careful study on false positive and false negative classes, following observations can be made.\\1) Instances with a large number of protein mentions in a single sentence can cause misclassification. For example, the maximum number of proteins in any instances of BioInfer and HPRD50 datasets are 26 and 24, respectively. These large number of proteins present in a single instance may lead the network to misclassificaton.\\
    2) Few samples contain repeated mentions of the same protein. This adds noise and might lead to losing useful contextual information.\\
   3) To get a consistent graph from molecular structure, the nodes were required to be of the same length. This is done by padding the vectors with zeros, and when the PDB is not available, a null vector is used for consistency. A better handling of missing data will help in learning the proposed model. 
\section{Conclusion}\label{conclusion}
This work presents a novel modality-agnostic Graph-based framework to identify the interactions between the proteins. Specifically, we explored two modalities: \textit{textual}, and \textit{molecular structure} that enable the model to learn the domain-specific multi-omnics information complementary with the task-specific contextual information.
A detailed comparative results and analysis proves that our proposed multi-modal approach can capture underlying molecular structure information without relying on sophisticated modality-specific architectures.
Future work aims at extending this study to the other related tasks like drug-drug interactions.  
\section*{Acknowledgment}
Sriparna Saha gratefully acknowledges the Young Faculty Research Fellowship (YFRF) Award, supported by Visvesvaraya Ph.D. Scheme for Electronics and IT, Ministry of Electronics and Information Technology (MeitY), Government of India, being implemented by Digital India Corporation (Formerly Media Lab Asia) for carrying out this research. 



\bibliographystyle{acl_natbib}
\bibliography{anthology,acl2021}


\end{document}